\documentclass[10pt,conference]{IEEEtran}
\IEEEoverridecommandlockouts
\usepackage{cite}
\usepackage{amsmath,amssymb,amsfonts}
\usepackage{algorithmic}
\usepackage{graphicx}   
\usepackage{booktabs}   
\usepackage{multirow}   
\usepackage{xcolor}     
\usepackage{graphicx}
\usepackage{textcomp}
\usepackage{pifont}
\usepackage{xspace}
\usepackage{lineno,hyperref}
\usepackage[utf8]{inputenc}
\def\BibTeX{{\rm B\kern-.05em{\sc i\kern-.025em b}\kern-.08em
    T\kern-.1667em\lower.7ex\hbox{E}\kern-.125emX}}

\usepackage[table]{xcolor}
\newcommand{\greenum}[1]{\cellcolor[HTML]{E6F8E0}{\textcolor{black}{#1}}}
\newcommand{\bluenum}[1]{\cellcolor[HTML]{D4E6F1}{\textcolor{black}{#1}}}
\newcommand{\bluetext}[1]{\colorbox[HTML]{D4E6F1}{#1}}
\newcommand{\greentext}[1]{\colorbox[HTML]{E6F8E0}{#1}}

\newcommand{\tabincell}[2]{\begin{tabular}{@{}#1@{}}#2\end{tabular}}
\newcommand{\xmarkg}{\textcolor{green}{\ding{55}}\xspace}%
\newcommand{\cmarkg}{\textcolor{red}{\ding{51}}\xspace}%

\usepackage{pifont}
\usepackage{xcolor}

\begin{document}

\title{STEDiff: Strengthening Text Embedding for Text-to-Image Alignment in Diffusion Model
}
\author{
\IEEEauthorblockN{
Hailan Zhang$^1$, Haipeng Liu$^{1,*}$\thanks{*Haipeng Liu is the corresponding author},
Bo Fu$^2$, Yang Wang$^1$
}

\IEEEauthorblockA{
\textit{$^1$ School of Computer Science and Information Engineering, Hefei University of Technology, Hefei, China}\\
\textit{$^2$ School of Computer and Artificial Intelligence, Liaoning Normal University, Dalian, China}\\
zhhl@mail.hfut.edu.cn,
hpliu\_hfut@hotmail.com,
fubo@lnnu.edu.cn,
yangwang@hfut.edu.cn
}
}



\maketitle

\begin{abstract}
Although pretrained text-to-image (T2I) generation models can produce high-quality images, they often fail to faithfully reflect the semantic intent of complex prompts due to stochastic noise and inherent model limitations. This issue frequently manifests as the model overlooking specific objects or failing to correctly bind attributes to their corresponding entities—a challenge referred to as \textit{semantic alignment}. Unlike existing approaches that rely on computationally expensive fine-tuning or labor-intensive layout priors, we propose \textit{STEDiff}, a training-free method designed to enhance semantic representations directly within the \textit{text-embedding space}. Specifically, we introduce a method that primarily leverages [EOT] tokens to strengthen the relevant semantics of sub-sentences, and then replaces the corresponding tokens in the original prompt. Furthermore, a novel semantic enhancement loss is incorporated to enforce spatial constraints, ensuring that the semantics of each entity are precisely mapped to their respective image regions. Extensive quantitative and qualitative evaluations on the T2I-CompBench demonstrate that our method notably improves semantic consistency and generation integrity in complex scenarios.
\end{abstract}

\section{Introduction}
\begin{figure}[htbp]
    \centering
    \includegraphics[width=\columnwidth]{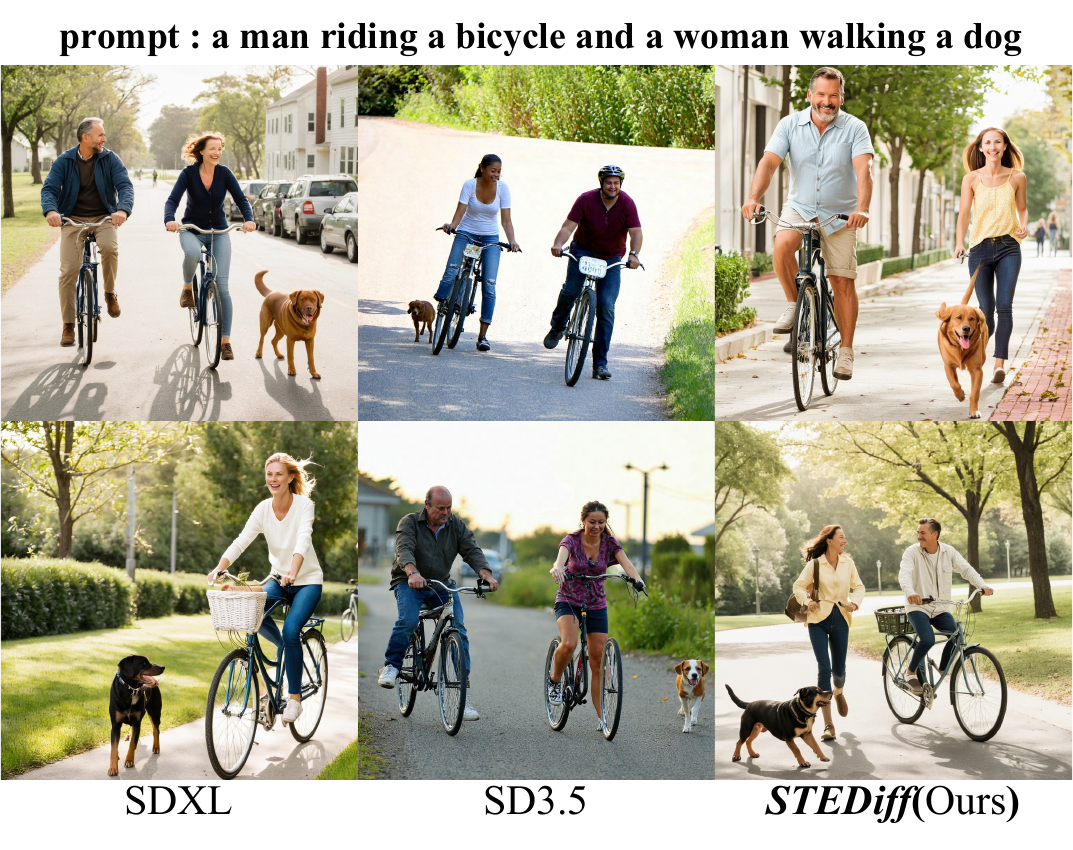}
    \caption{\textbf{Comparison of Different Methods.} Facing complex prompts, T2I models often suffer from semantic binding issues. For example, attributes associated with ``woman'' may be incorrectly bound to ``bicycle'', or the ``man'' entity may fail to be generated correctly. To address these challenges, we propose the \textit{STEDiff} method.}
    \label{fig1}
\end{figure}

In recent years, T2I diffusion models\cite{balaji2022ediff, ramesh2021zero,saharia2022photorealistic,ramesh2022hierarchical,liu2025ones,liu2024structure,wang2022progressive} have led to remarkable improvements in generating high-quality, detail-rich images from text prompts. These models are usually trained on large text-image datasets using contrastive loss to establish alignment between different spaces. However, due to the limitations of the dataset and insufficient training, aligning the generated images with the prompts, which is referred to as \textit{semantic alignment}\cite{li2023divide}, remains a significant challenge. Typical issues include: 1) binding objects with their attributes(\textit{attribute binding}); 2) binding objects with their sub-objects(\textit{sub-objects binding}); and 3) the inability to generate noun phrases from the prompts(\textit{noun missing}). For example, as shown in Fig.~\ref{fig1}, even the state-of-the-art model such as stable-diffusion-xl-base-1.0\cite{podell2023sdxl}, Stable Diffusion 3.5\cite{esser2024scaling} struggles to properly handle the relationships between different terms in the text prompt, leading to issues of text-image inconsistency.

To improve the alignment between image and text during the generation process, several methods have been proposed. Among these approaches, some focus on optimizing latent representations\cite{rassin2023linguistic, ge2023expressive,qian2022switchable,qian2023adaptive,wang2024unpacking} to enhance consistency or fine-tuning T2I models\cite{jiang2024comat,hu2024ella,feng2024ranni,liu2025few,chen2024pixart}. In parallel, others introduce additional constraints, such as layout guidance\cite{xie2023boxdiff, phung2024grounded}, or incorporate Large Language Models (LLMs)\cite{lian2023llm, gani2023llm} to improve the model's prompt comprehension. Furthermore, several methods\cite{qian2023rethinking, hu2024token, chen2024cat,liu2022delving} prioritize the optimization of text embeddings to strengthen semantic binding. However, these methods still encounter limitations. For example, \cite{jiang2024comat} requires substantial computational resources, \cite{xie2023boxdiff} sacrifices the diversity of generated images, and \cite{hu2024token} performs simple token summation, which leads to low coherence in the images.

In this paper, we rethink the role of the text embeddings generated by the CLIP text encoder. Specifically, we perform singular value decomposition on the text embeddings to obtain singular values that represent the primary semantic components of the prompt. Building on the premise that noun phrases constitute the core semantic backbone, we enhance these singular values to effectively mitigate the neglect of such entities during generation. Meanwhile, due to the causal nature of the text encoder, tokens with higher indices—culminating in the [EOT] symbol—accumulate the most comprehensive contextual information. By leveraging the [EOT] token as a global semantic anchor and enhancing the singular values, our method ensures strong semantic binding, as demonstrated by extensive empirical and theoretical evidence.

Based on the above observations, we propose a text embedding-based semantic activation to assist the model in better understanding text prompt. Although many current methods aim to address the issue of attribute binding, there has been relatively little exploration into resolving the sub-objects binding\cite{hu2024token}. As an illustration, consider the prompt ``a man riding a bicycle and a woman walking a dog'', the man is expected to ride a bicycle rather than walk a dog. In order to address the above challenges, we propose \textit{STEDiff}, a novel train-free image generation alignment method, it first split the complex prompts by different entity-nouns, such as \textit{man} and \textit{woman}, thereby obtaining the sub-sentences ``\textit{a man riding a bicycle}'' and ``\textit{a woman walking a dog}'', represented as man* and woman*, then strength the semantics of the sub-sentence separately, finally perform the replacements from the initial complex prompts. Instead of operating on the complete prompt, we treat objects and their sub-objects as a whole, avoiding semantic leakage issues from other entities. Using woman* as an example, we build a text embedding matrix that includes both the intended content(woman walking a dog) and end-of-token [EOT] embeddings, applying \textit{soft-weighted regularization} to this matrix can further activate the target semantics. 

Meanwhile, since the overall layout is largely determined at the early stages of the diffusion process, we optimize the text embeddings at inference-time using a semantic enhancement loss and an entropy loss. The semantic enhancement loss strengthens the associations between objects and their sub-objects, preventing certain objects from being omitted during generation, thereby improving the alignment among textual concepts and consequently enhancing the structural integrity of the generated images. The entropy loss encourages each prompt token to attend to its corresponding region, thereby mitigating semantic leakage and preventing sub-objects from being incorrectly bound.

To evaluate the effectiveness of \textit{STEDiff}, we perform a quantitative analysis on the widely adopted T2I-CompBench benchmark\cite{huang2023t2i}. Through a comparative evaluation across various baselines, datasets, and metrics, it is clear that \textit{STEDiff} outperforms the others, particularly in scenarios involving multi-object and multi-attribute generation. Notably, \textit{STEDiff} is train-free and user-friendly, primarily modifying in the text embedding space without requiring additional input conditions, such as layout information, which further highlights the superiority of our approach. Overall, the main contributions of our work are as follows:
\begin{itemize}
    \item We conduct a comprehensive analysis of the text embeddings, highlighting that the [EOT] token contains redundant semantic information(in Fig.~\ref{fig4}). By enhancing the semantic information of [EOT], we can improve the semantic features in generated images(in Fig.~\ref{fig6}).
\end{itemize}
\begin{itemize}
    \item We propose \textit{STEDiff}(Fig.~\ref{fig2}), a train-free method that improves the semantic alignment capability of existing T2I models when dealing with complex prompts, through simple enhancement and replacement of text embeddings.
\end{itemize}
\begin{itemize}
    \item We conducted extensive experiments(Fig.~\ref{fig7}, Tab.~\ref{tab:bvqa}) to demonstrate the effectiveness of our method compared to current SOTA T2I models.
\end{itemize}

\section{Related Work}

\subsection{Text-to-Image Diffusion Model} 
Diffusion models, such as Stable Diffusion\cite{rombach2022high}, DALL-E 2\cite{ramesh2022hierarchical}, and Imagen\cite{saharia2022photorealistic}, have emerged as the leading approach\cite{dhariwal2021diffusion, balaji2022ediff, esser2024scaling, labs2025flux1kontextflowmatching, podell2023sdxl, ramesh2021zero, rombach2022high} in T2I generation area. These models operate as stochastic diffusion processes\cite{ho2020denoising}, beginning with a simple distribution \(\boldsymbol{x}_T\sim\mathcal{N}(\boldsymbol{0},\boldsymbol{I}),\) and gradually adding noise through the forward process, then removes noise over \textit{T} steps using a denoising network \(\boldsymbol{\epsilon}_{\boldsymbol{\theta}}\) to generate samples in the reverse process. By injecting ``external information'' such as text embeddings \({\boldsymbol{p}}\) obtained by a text encoder such as CLIP\cite{radford2021learning} into the iterative denoising process, the model can be guided\cite{ho2022classifier} to generate the desired content by \(\boldsymbol{\epsilon}_{\boldsymbol{\theta}}(\boldsymbol{x}_t,\boldsymbol{p},t)\). 

\begin{figure*}[htbp]
    \centering
    \includegraphics[width=\textwidth]{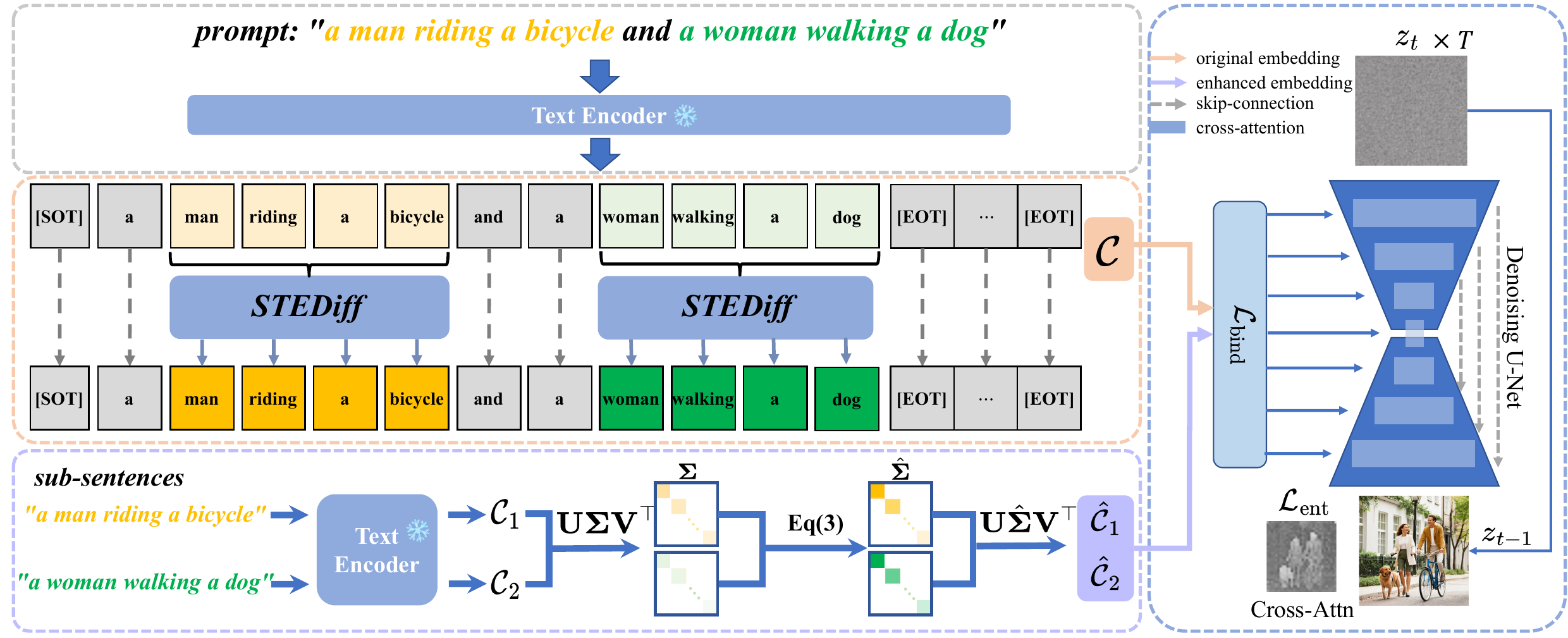}
    \caption{\textbf{Overview of \textit{STEDiff}}. Our method starts by splitting complex prompts and treating each sub-sentence as a clean, prompt-level supervision signal. During the denoising process, we apply \textit{\textbf{STEDiff}} to the resulting sub-sentences to obtain enhanced embeddings for improved image feature representation. In the early stages of denoising, the $\mathcal{L}_{\mathrm{bind}}$ and $\mathcal{L}_{\mathrm{ent}}$ are used in tandem to update and replace tokens between the original and enhanced embeddings.}
    \label{fig2}
\end{figure*}

\subsection{Text-to-Image Alignment}
To improve T2I alignment in diffusion models, numerous studies~\cite{dahary2024yourself,wang2024phased,shen2024rethinking,tumanyan2023plug} have been proposed. Layout-based methods leverage predefined layouts from users~\cite{xie2023boxdiff,phung2024grounded,wang2024compositional,wang2024instancediffusion} or LLMs~\cite{yang2024mastering,li2024mulan,lian2023llm,gani2023llm} to guide generation within specified regions. Other approaches~\cite{chefer2023attend,li2023divide,rassin2023linguistic,ge2023expressive} employ test-time optimization to align cross-attention maps between attributes and subjects. Fine-tuning-based methods~\cite{jiang2024comat,hu2024ella,feng2024ranni,chen2024pixart} mitigate semantic incoherence by retraining models on large-scale paired datasets. However, these methods often suffer from limited diversity, difficulties with multi-object or multi-attribute prompts, or substantial computational overhead.

\subsection{Text Embedding Approaches for Text-to-Image}
Recent some research has focused on the text embedding space in T2I tasks\cite{seo2025geometrical,yu2024uncovering,wu2025core}. \cite{chen2024cat} introduced that the causal processing of the CLIP text encoder leads to the accumulation of text information, resulting in biases and losses in the generated images. \cite{li2024get} suppresses the generation of certain text information by processing the [EOT] information within the text embeddings. Magnet addresses attribute bias in the text embedding space by introducing positive and negative binding vectors and highlights the contextual issues in padding embeddings that entangle different concepts. ToMe\cite{hu2024token} leverages the additivity of text embeddings to incorporate all attributes into the objects token. 

\section{Method}
We aim to enhance semantic coherence in T2I tasks within diffusion models. By analyzing the text embeddings, we found that in addition to the object token, [EOT] tokens accumulate a significant amount of semantic information that aids in image generation(in Fig.~\ref{fig4}). By activating the target semantic information, we can facilitate the correct alignment of each target objects with their attributes. In this section, we first introduce some preliminary concepts in diffusion models(Sec.\ref{3section:A}). Subsequently, we analyze the importance of text embeddings in the generation process through a series of experiments and introduce our motivation(Sec.\ref{3section:B}). Finally, we provide a detailed explanation of the specifics of the semantic enhancement techniques and semantic enhancement loss of our method(Sec.\ref{3section:C}). Fig.~\ref{fig2} presents an illustration of our \textit{STEDiff} method.

\subsection{Preliminary}
\label{3section:A}
Our method is implemented using stable diffusion(SD) model, which belongs to the family of latent diffusion models, it usually includes a text encoder, a variational autoencoder(i.e.,a encoder \(\mathcal{E}\) and a decoder \(\mathcal{D}\)), and a denoising UNet(i.e.,\(\epsilon_{\theta}\) with parameter \(\theta\)). In the forward process, noise \(\epsilon\) is added to the original latent representation $z_0$. Then a denoising function \(\epsilon_{\theta}\) is trained to predict the noise \(\epsilon\)  added to $z_0$, following the objective:
\begin{equation}
L_{LDM}:=\mathbb{E}_{z_0,\epsilon\thicksim\mathcal{N}(0,1),t\thicksim(1,T)}{\left[\|\epsilon-\epsilon_\theta(z_t,t,\tau_\xi(\mathcal{P}))\|_2^2\right]}
\end{equation}
where $\tau_{\xi}$ is a pre-trained CLIP text encoder, $\mathcal{P}$ is the text prompt, and $z_t$ is a noisy latent sample at timestep $t \sim [1, T]$. The final decoder reverses the latent representation into an image $\hat{x} = \mathcal{D}(z_0)$.

By introducing the text embeddings obtained from the CLIP text encoder, the SD model incorporates cross-attention layers. Cross-attention maps can be extracted internally from \(\epsilon_\theta(z_t,t,\tau_\xi(\mathcal{P})\) as high-dimensional tensors, ensuring that the generated images are consistent with the text prompts. The \({i}\)-th cross-attention map \(A^{(i)}\in\mathbb{R}^{h\times w}\) is computed as
\begin{equation}
A^{(i)}=\mathrm{Softmax}(\frac{Q^{(i)}(K^{(i)})^T}{\sqrt{d}})
\end{equation}
where \(\sqrt{d}\) is a scaling factor, the features map of size is \(h\times w\). By applying a linear transformation, the model converts the latent and text embeddings into queries \({Q}\) and keys \({K}\) for computing attention weights and identifying the most relevant information. 

\begin{figure}[htbp]
    \centering
    \includegraphics[width=\columnwidth]{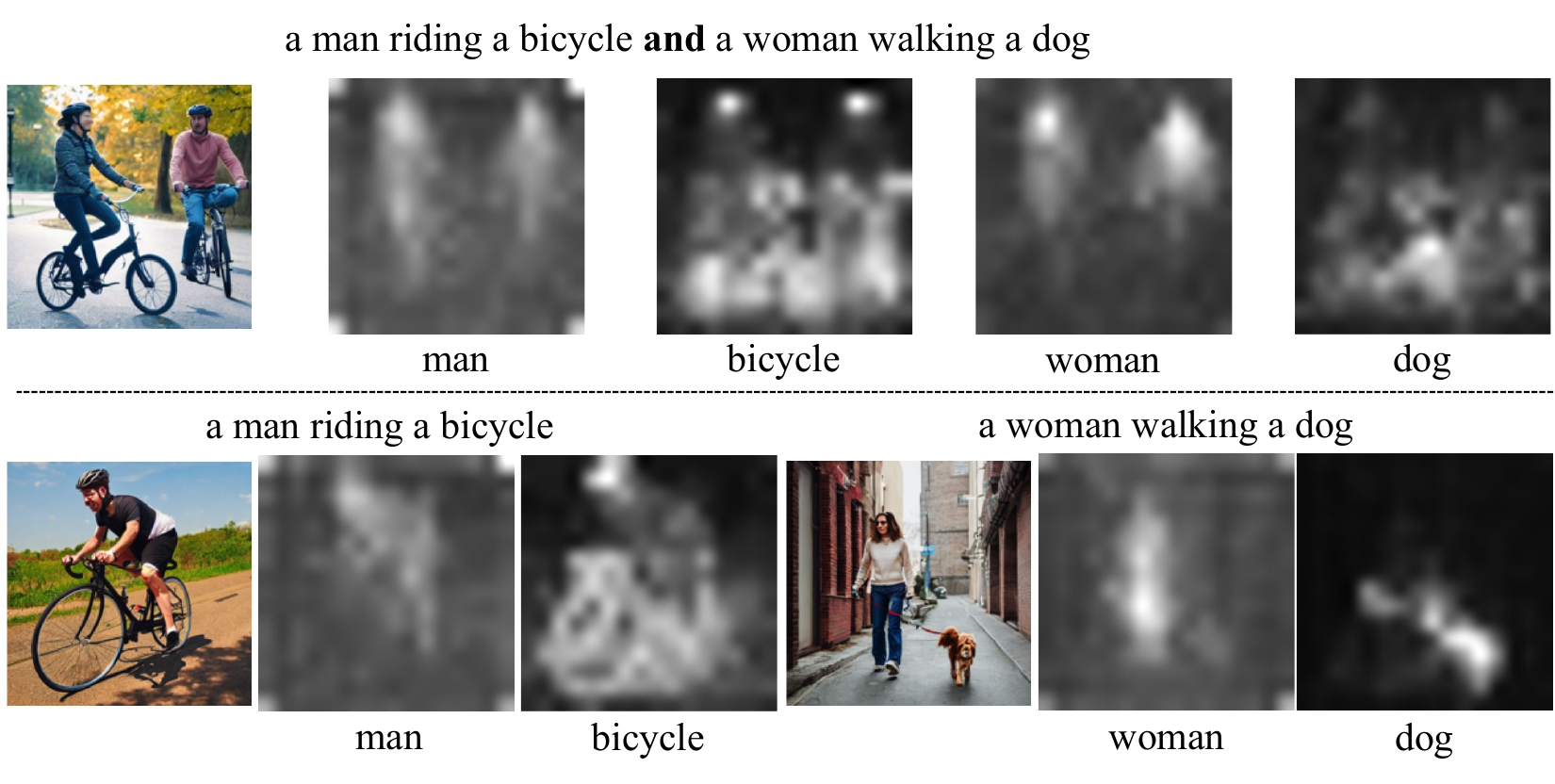}
    \caption{\textbf{Attention visualization.} The cross-attention map for each token of the prompt ``a man riding a bicycle and a woman walking a dog'' and its sub-sentences is visualized. When faced with a complex prompt, there is a noticeable entanglement between different subjects, whereas simple sub-sentences do not exhibit this behavior.}
    \label{fig3}
\end{figure}

After visualizing the extracted cross-attention maps, Fig.~\ref{fig3} observes that each token within the prompt corresponds to its own attention map. However, when the prompt contains multiple objects or attributes, the attention distributions of different subjects become intertwined, showing obvious entanglement, which is the main cause of the \textit{semantic alignment} issue.

\subsection{Analysis of Text Embeddings}
\label{3section:B}
\begin{figure}[htbp]
    \centering
    \includegraphics[width=\columnwidth]{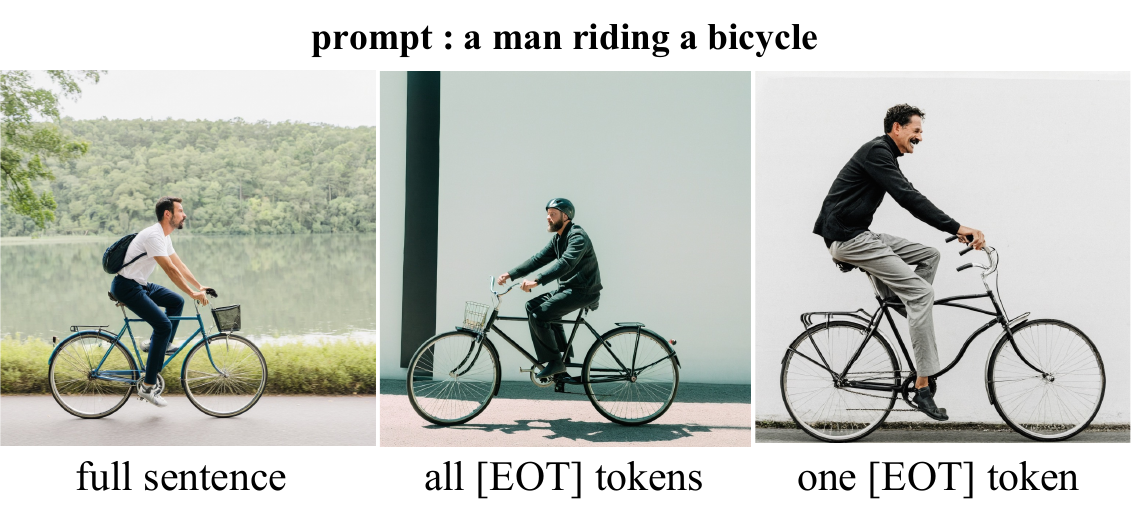}
    \caption{\textbf{Analysis of Text Embeddings.} [EOT] tokens contain a significant amount of information, and even a single [EOT] token can generate images with semantically similar content to those produced by the full prompt.}
    \label{fig4}
\end{figure}

As one of the most critical components in the SD, the CLIP text encoder \(\tau_{\xi}\) transforms the input prompt into a fixed-dimensional text embedding. For a given text prompt \(\mathcal{P}\), the \(\tau_{\xi}\) tokenizes it by adding a start token [SOT] at the beginning and [EOT] at the end, while padding with additional [EOT] tokens as necessary to extend it to a fixed length of $N$, where $N = 77$ in our experiments. For example, when the prompt is ``a man riding a bicycle'', it is first tokenized and then processed by the \(\tau_{\xi}\) into a sequence such as \(\boldsymbol{c}=\{\boldsymbol{c}^{SOT},\boldsymbol{c}_0^a,\boldsymbol{c}_1^{man},\boldsymbol{c}_2^{riding},\boldsymbol{c}_3^{a},\boldsymbol{c}_4^{bicycle},\boldsymbol{c}_{|\boldsymbol{p}|}^{EOT},\cdots,\boldsymbol{c}_{N\boldsymbol{-}|\boldsymbol{p}|\boldsymbol{-}2}^{EOT}\}\). Due to the causal manner of the CLIP text encoder, each token can only ``see'' the preceding tokens. Consequently, the \(\boldsymbol{n}^{th}\) token becomes entangled with the previous \(\boldsymbol{n}-1\) tokens, causing the later tokens to carry a substantial amount of accumulated semantic information. Through observation, we find that similar images can be generated not only when all [EOT] tokens(\(\{\boldsymbol{c}_{|\boldsymbol{p}|}^{EOT},\cdots,\boldsymbol{c}_{N\boldsymbol{-}|\boldsymbol{p}|\boldsymbol{-}2}^{EOT}\}\)) are used, but even when only a single [EOT] token is provided. Fig.~\ref{fig4} indicates that a single [EOT] token has already accumulated a substantial amount of semantic information.
\subsection{Analysis of [EOT] tokens}
\label{3section:C}

To further reveal the issues that arise when the prompt contains multiple subject attributes, we calculated the information entropy of each token in the prompt and found that tokens appearing earlier in the sequence tend to exhibit higher entropy, as illustrated in Fig.~\ref{fig5}(a). As the token index increases, the entropy sharply decreases at the [EOT] position of each sentence, after which the information entropy of subsequent tokens gradually diminishes. Early tokens dominate the overall semantic representation, whereas the [EOT] token as a semantic boundary, thereby revealing an inherent limitation of the text encoder in modeling prompts.

Due to the padding embedding mechanism, embeddings at later indices gradually deviate from the true semantic space, thereby leading to semantic degradation. Furthermore, As shown in Fig.~\ref{fig5}(b), we compute the cosine similarity between the first [EOT] token and the subsequent padded [EOT] tokens under both the original prompts and their corresponding sub-sentences. The curve drops much more sharply for complex concepts than for simple ones, suggesting that, in complex clauses, the subsequent padded embeddings are more prone to forgetting part of the contextual information retained in the first [EOT].

However, prior studies\cite{feng2022training,li2024get} have shown that these padded embeddings are crucial for the quality of image generation, and that the paddings are entangled with one another, making it impossible to manipulate any single concept in isolation.

\subsection{STE: Strength Text Embedding }
As observed in the preceding section, the SD model is more prone to forgetting when faced with complex prompts, and the [EOT] token contains significant semantic information. To mitigate this issue, we propose strengthening the text–image alignment by enhancing the embedding of the [EOT] tokens.

\label{section:D}
\subsubsection{Text Embedding-Based Semantic Activation in diffusion model}
\label{sec:semantic-activation}
Inspired by \cite{gu2014weighted, li2024get}, we analyze the text-embedding matrix of each sub-sentence, under the assumption that its dominant singular values represent the fundamental information of the prompt. Taking the sub-sentence ``a man riding a bicycle'' as an illustrative example, we apply SVD to its text-embedding matrix, i.e., \(\boldsymbol{c}=\mathbf{U}\boldsymbol{\Sigma}\mathbf{V}^\top\), where \(\boldsymbol{\Sigma}=diag(\sigma_0,\sigma_1,\cdots,\sigma_{n})\), and \(\sigma_0\geq\cdots\geq\sigma_{n}\).

To enhance the representation, we introduce a set of amplification coefficients for the singular values to strengthen its expressiveness. The formulation is given by:
\begin{equation}
\hat{\sigma}=\beta e^{\alpha\sigma}*\sigma
\end{equation}

where $e$ denotes the base of the natural exponential function, and $\alpha$ and $\beta$ are positive scalar parameters.
After obtaining the updated $\hat{\boldsymbol{\Sigma}} = \operatorname{diag}(\hat{\sigma}_0, \hat{\sigma}_1, \cdots, \hat{\sigma}_n)$, 
the enhanced text-embedding matrix $\hat{\boldsymbol{c}} = \mathbf{U}\hat{\boldsymbol{\Sigma}}\mathbf{V}^\top$ can be reconstructed accordingly.

After obtaining the enhanced text embeddings, we generate images conditioned on these embeddings and then extract visual features using the DINOv3 model. The extracted features are subsequently projected into a 2D space via PCA for dimensionality reduction and visualization, as illustrated in Fig.~\ref{fig6}(a). This is further depicted in Fig.~\ref{fig6}(b), this pipeline yields richer and more informative visual representations, making it easier for the model to retain and reflect the information encoded in the text embeddings.

\subsubsection{Semantic Bind Loss} 
As stated in Section~\ref{sec:semantic-activation}, enhancing semantic information enables the generation of images with richer and more discriminative features. When dealing with complex prompts $\mathcal{P}$ of  ``a man riding a bicycle and a woman walking a dog''. Let $\mathcal{C}=\{\boldsymbol{c}^{\mathrm{SOT}}, \boldsymbol{c}_0^{a}, \boldsymbol{c}_1^{\mathrm{man}}, \ldots,
\boldsymbol{c}_7^{\mathrm{woman}}, \boldsymbol{c}_{\lvert \boldsymbol{p} \rvert}^{\mathrm{EOT}}, \ldots,
\boldsymbol{c}_{N-\lvert \boldsymbol{p} \rvert-2}^{\mathrm{EOT}}\} = \{ \boldsymbol{{C}^{man}},\boldsymbol{{C}^{woman}}\}$. It is necessary to strengthen the association between each sub-sentence's object and its corresponding sub-object to prevent incorrect bindings.

To this end, we treat each sub-sentence separately, i.e., $\mathcal{C}_1=\{\boldsymbol{c}^{\mathrm{SOT}}, \boldsymbol{c}_0^{a}, \boldsymbol{c}_1^{\mathrm{man}}, \ldots,
\boldsymbol{c}_4^{\mathrm{bicycle}}, \boldsymbol{c}_{\lvert \boldsymbol{p}_1 \rvert}^{\mathrm{EOT}}, \ldots,
\boldsymbol{c}_{N-\lvert \boldsymbol{p}_1 \rvert-2}^{\mathrm{EOT}}\}$
and
$\mathcal{C}_2=\{\boldsymbol{c}^{\mathrm{SOT}}, \boldsymbol{c}_0^{a}, \boldsymbol{c}_1^{\mathrm{woman}}, \ldots,\boldsymbol{c}_4^{\mathrm{dog}},
\boldsymbol{c}_{\lvert \boldsymbol{p}_2 \rvert}^{\mathrm{EOT}}, \ldots,
\boldsymbol{c}_{N-\lvert \boldsymbol{p}_2 \rvert-2}^{\mathrm{EOT}}\}$, we applied the \textit{STEDiff} to enhance the text embeddings of $\mathcal{C}_1$ and $\mathcal{C}_2$, respectively, thereby obtaining the enhanced representations of $\hat{\mathcal{C}}_1 = \{\boldsymbol{\hat{C}}^{man}\}
$ and $\hat{\mathcal{C}_2} = \{\boldsymbol{\hat{C}}^{woman}\}$. 

We treat each sub-sentence as a clean prompt-level supervision signal, with the aim of making each component of $\mathcal{P}$ match its corresponding sub-sentence as closely as possible. In other words, the objective is to make $\boldsymbol{\hat{C}^{man}}$ as close as possible to $\boldsymbol{{C}^{man}}$ in the text embedding space, and the same applies to woman. Here, we aim to encourage the binding between each subject and its corresponding attributes. To this end, we introduce a positive binding loss:

\begin{figure}[htbp]
    \centering
    \includegraphics[width=\columnwidth]{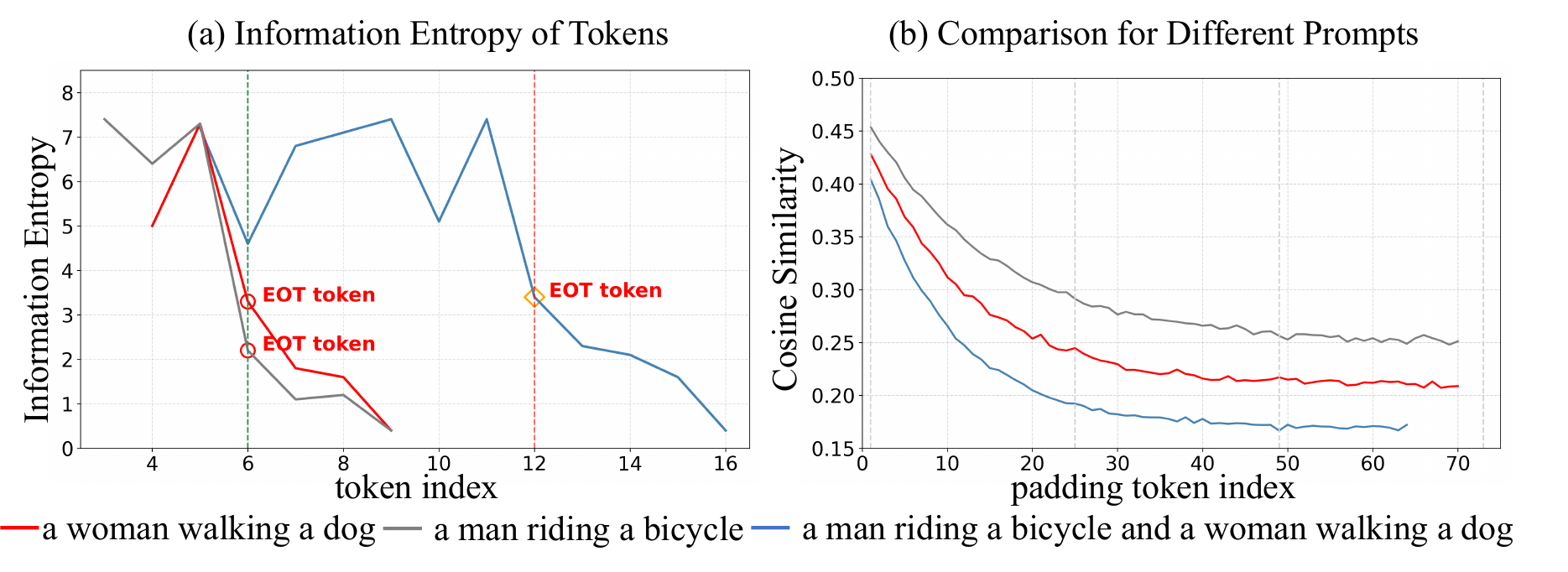}
    \caption{\textbf{Analysis of [EOT] tokens.} (a)Earlier tokens exhibit higher information entropy, which sharply decreases upon reaching the first [EOT] token as the token index increases. (b)When faced with complex prompts, the text embeddings are more likely to deviate from the true semantic space.}
    \label{fig5}
\end{figure}

\begin{equation}
\mathcal{L}_{\mathrm{bind}}
= \sum_{k=1}^{K}
\left\|
\epsilon_\theta(z_t, \hat{C}_k, t)
-
\epsilon_\theta(z_t, {C}_k, t)
\right\|_2^2 .
\end{equation}

As illustrated in Fig.~\ref{fig3}, each token in the textual prompt contains a substantial amount of information, which may easily lead to semantic leakage. To enhance the concentration of attention, inspired by \cite{hu2024token, shannon1948mathematical}, we combine the previously defined loss term with an entropy-based loss to construct the overall loss function $\mathcal{L}_{\mathrm{sel}}$. The entropy-based loss $\mathcal{L}_{\mathrm{ent}}$ is defined as:
\begin{equation}
\mathcal{L}_{\mathrm{ent}}=-\sum_{k\in{K}}\sum_{p_i\in \mathcal{A}_{k}}p_i\log p_i,
\end{equation}
where $\mathcal{A}_{k}$ represents the cross-attention map associated with the $k$-th token, and $p_i$ denotes the attention probability at the $i$-th spatial position in $\mathcal{A}_{k}$.

By combining the loss functions $\mathcal{L}_{\mathrm{bind}}$ and $\mathcal{L}_{\mathrm{ent}}$, the final loss function $\mathcal{L}_{\mathrm{sel}}$ is defined as:
\begin{equation}
\mathcal{L}_{\mathrm{sel}}=\mathcal{L}_{\mathrm{bind}}+\lambda\mathcal{L}_{\mathrm{ent}}
\end{equation}
where $\lambda$ is a trade-off hyperparameter. During the early diffusion steps $t$, we update the latent vectors via gradient descent:
\begin{equation}
\mathbf{z}_{t}^{\prime} \leftarrow \mathbf{z}_{t} - \eta \cdot \nabla_{\mathbf{z}_{t}} \mathcal{L}_{\mathrm{sel}}
\end{equation}
where $\eta$ is the step size.

It is worth noting that existing studies have not explicitly demonstrated that activating semantic information can enhance the binding between subjects and attributes. During optimization, for the text embedding of each subject, we apply \textit{STEDiff} to emphasize its corresponding semantics.

\section{Experiments}
\subsection{Evaluation Benchmarks and Baseline Methods}
\label{section:A}
We carry out extensive experiments on T2I-CompBench\cite{huang2023t2i}, a widely used benchmark designed to evaluate text-to-image generation under complex compositional prompts. The benchmark offers a broad and diverse collection of prompts for comprehensive performance assessment. Specifically, we focus on the color, shape, and texture categories to measure the model’s attribute-binding capability. In addition, we employ ImageReward \cite{xu2023imagerewardlearningevaluatinghuman} to further examine the perceptual quality of the generated images and to estimate human preference scores. As the first general-purpose human-preference reward model for text-to-image generation, ImageReward supports reliable scoring and ranking of generated outputs. We follow the evaluation protocol\cite{chen2025detail++} using the BLIP-VQA score and human preference as evaluation metrics.

\begin{figure}[htbp]
    \centering
    \includegraphics[width=0.7\columnwidth,trim=10 10 10 0,clip]{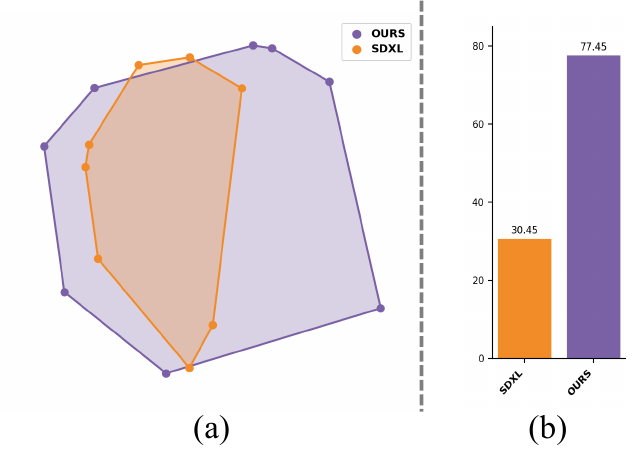}
    \caption{\textbf{Comparison of image features.} (a)Image features are extracted using DINOv3 and visualized through PCA reduction. Compared to the SDXL, our method demonstrates more diverse and rich image features, which contribute to enhanced semantic binding. (b)Statistics of the average feature area calculated from the dimensionality-reduced areas demonstrate that our method reveals more image features.}
    \label{fig6}
\end{figure}

To further validate the effectiveness of the proposed \textit{STEDiff} method, we compare it with a variety of baseline approaches: ToMe\cite{hu2024token}, ELLA\cite{hu2024ella}, Attention Regulation\cite{zhang2024enhancingsemanticfidelitytexttoimage}, Syngen\cite{rassin2024linguisticbindingdiffusionmodels}, Attend and Excite\cite{chefer2023attend}, PixArt\cite{chen2024pixart} and SDXL\cite{podell2023sdxl}. For each prompt, we employ the NLP toolkit SpaCy\cite{honnibal2017spacy} to identify noun-related subjects. Meanwhile, $\alpha = 0.004$ and $\beta = 1.1$ in our experiments. All experiments were run on an Nvidia A40 GPU.

\begin{figure*}[t]
    \centering
    \includegraphics[width=\textwidth]{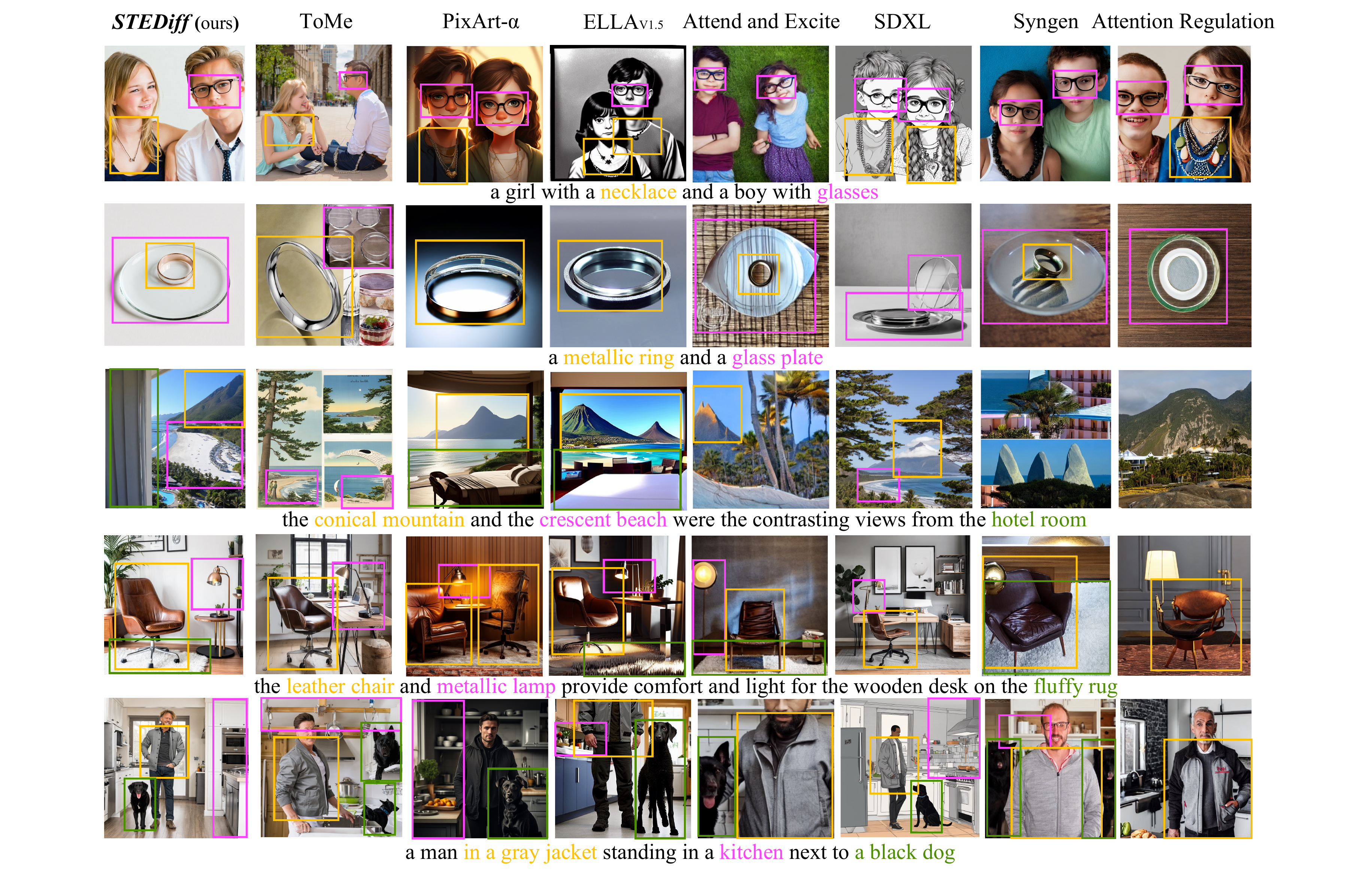}
    \caption{\textbf{Qualitative comparison of \textit{STEDiff} against semantic binding baselines in different scenarios.} Whether handling objects and their attributes(second row) or sub-objects(first row), \textit{STEDiff} consistently maintains high alignment. Moreover, when faced with complex prompts, it reliably generates images that align accurately with the text description.}
    \label{fig7}
\end{figure*}

\subsection{Quantitative Comparison}
\label{section:B}
As summarized in Tab.~\ref{tab:bvqa}, based on extensive quantitative evaluation against diverse semantic binding methods, \textit{STEDiff} attains consistently superior or comparable BLIP-VQA scores on the color, shape, and texture subsets, highlighting its capability to alleviate attribute confusion. Meanwhile, \textit{STEDiff} achieves favorable human-preference scores on ImageReward\cite{xu2023imagerewardlearningevaluatinghuman}, which serve as a proxy for real human preferences, demonstrating stronger alignment with human judgments. Overall, our method demonstrates competitive performance and, on average, outperforms the existing baseline approaches.

\begin{table*}[t]
\centering
\vspace{-5mm}
\renewcommand{\arraystretch}{1}  
\setlength{\tabcolsep}{8pt}  
\caption{Quantitative results for semantic binding assessment on various benchmarking subsets. We denote the best score in \bluetext{blue}, and the second-best score in \greentext{green}.}
\label{tab:bvqa}
\tiny 
\resizebox{\textwidth}{!}{%
\begin{tabular}{cc | ccc | ccc}
\toprule
\multirow{2}{*}{Method} &
\multirow{2}{*}{\tabincell{c}{Train}} &
\multicolumn{3}{c|}{BLIP-VQA $\uparrow$} &
\multicolumn{3}{c}{Human-preference $\uparrow$} \\
& & Color & Texture & Shape & Color & Texture & Shape \\
\midrule
SD1.5\cite{rombach2022high} & \cmarkg & 0.4719 & 0.4334 & 0.3898 & -0.3600 & -0.689 & -0.483 \\
ELLA$_{1.5}$\cite{hu2024ella} & \cmarkg & \greenum{0.6911} & 0.6308 & 0.4938 & - & - & - \\
CoMat$_{1.5}$\cite{jiang2024comat} & \cmarkg & 0.6561 & 0.6190 & 0.4975 & - & - & - \\
SynGen$_{1.5}$\cite{rassin2024linguisticbindingdiffusionmodels} & \xmarkg & 0.6619 & 0.6451 & 0.4661 & 0.4326 & \greenum{0.5072} & 0.0426 \\
SDXL\cite{podell2023sdxl} & \cmarkg & 0.6369 & 0.5637 & 0.5408 & \bluenum{0.7330} & 0.1240 & \greenum{1.184} \\
Attention Regulation\cite{zhang2024enhancingsemanticfidelitytexttoimage} & \xmarkg & 0.5860 & 0.5173 & 0.4672 & 0.2680 & -0.603 & 1.118 \\
PlayG-v2\cite{playground-v2} & \cmarkg & 0.6208 & 0.6125 & 0.5087 & - & - & - \\
Ranni$_{xl}$\cite{feng2024ranni} & \cmarkg & 0.6893 & 0.6325 & 0.4934 & - & - & - \\
PixArt\cite{chen2024pixart} & \cmarkg & 0.6886 & \bluenum{0.7044} & \greenum{0.5582} & 0.2420 & -0.788 & 0.1650 \\
ToMe\cite{hu2024token} & \xmarkg & 0.6583 & 0.6371 & 0.5517 & -0.0280 & -0.851 & 0.4540 \\
\midrule
\textit{\textbf{STEDiff}}(Ours) & \xmarkg & \bluenum{0.7010} & \greenum{0.6971} & \bluenum{0.6096} & \greenum{0.6232} & \bluenum{0.7215} & \bluenum{1.1285} \\
\bottomrule
\end{tabular}
}
\end{table*}

\subsection{Qualitative Comparison}
\label{section:B}
Meanwhile, to more intuitively demonstrate the effectiveness of \textit{STEDiff}, we qualitatively compare it with existing semantic binding methods across multiple text-to-image generation models. Fig.~\ref{fig7} presents qualitative comparison results between our method and other approaches. The first row shows the results on \textit{object binding}, while the second row presents the results on \textit{attribute binding}. For complex prompts (rows 3--5), we report the outputs produced by our method, demonstrating its effectiveness in handling challenging compositional instructions. The results clearly demonstrate that \textit{STEDiff} achieves superior semantic binding performance in both object binding and attribute binding scenarios involving two objects, as well as under complex prompting conditions.

Owing to its reliance on composite tokens and simple summation, ToMe~\cite{hu2024token} is prone to entity fragmentation. PixArt~\cite{chen2024pixart} achieves high image quality but fails to maintain semantic consistency in binding tasks. ELLA~\cite{hu2024ella} improves text understanding with LLMs yet struggles with complex prompts, while Attend-and-Excite~\cite{chefer2023attend} and SDXL~\cite{podell2023sdxl} are limited in handling attribute binding and often overlook prompt attributes. SynGen~\cite{rassin2024linguisticbindingdiffusionmodels} and Attention Regulation~\cite{zhang2024enhancingsemanticfidelitytexttoimage} focus on attribute binding but fail to generalize to object-level or multi-subject scenarios. In contrast, \textit{STEDiff} produces images with more coherent compositions and better alignment with the input prompts.

\begin{figure}[htbp]
    \centering
    \includegraphics[width=\columnwidth]{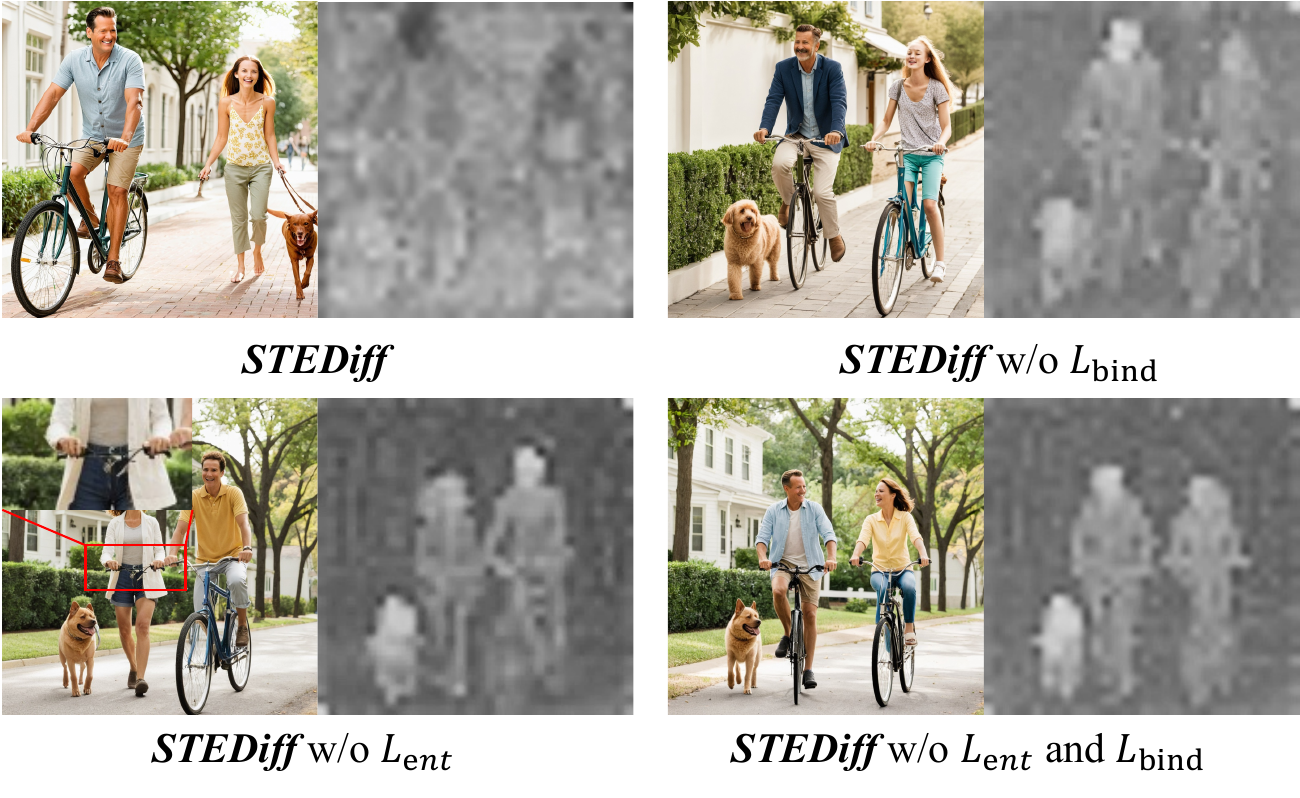}
    \caption{\textbf{Ablation study of different optimization terms in the attention mechanism.} Cross-attention visualization of the first [EOT] token shows that using any single optimization term alone does not yield ideal results.}
    \label{fig8}
\end{figure}

\subsection{Ablation Study}
\label{section:D}
We conduct ablation experiments to evaluate the effectiveness of the proposed components. The ablation study of each component is quantitatively reported in Tab.~\ref{tab:ablation_num}. We observe that using either $\mathcal{L}_{\mathrm{ent}}$ or $\mathcal{L}_{\mathrm{bind}}$ alone leads to only a slight improvement in performance, whereas combining them yields a significant performance gain. Moreover, for the same prompt, we generate one image under each configuration to enable a controlled comparison. Since our method performs activation and replacement by leveraging information from the [EOT] token, visualizing the cross-attention of the first few prompt tokens is not informative. The first [EOT] token often aggregates substantial semantic information accumulated from the preceding context; therefore, we visualize the cross-attention maps associated with this [EOT] token. As shown in Fig.~\ref{fig8}, incorporating our $\mathcal{L}_{\mathrm{bind}}$ enables the model to generate images that better match the prompt, although fine-grained details remain imperfect. In contrast, using the $\mathcal{L}_{\mathrm{ent}}$ alone is insufficient to accomplish the binding task.

\begin{table}[t]
\centering
\vspace{-5mm}  
\renewcommand{\arraystretch}{1}  
\setlength{\tabcolsep}{8pt}  
\caption{Ablation Study conducted on the T2I-CompBench benchmark.}
\resizebox{0.99\columnwidth}{!}{%
\begin{tabular}{ccc|ccc}
\toprule
\multirow{2}{*}{Conf.} &
\multirow{2}{*}{$\mathcal{L}_{bind}$} &
\multirow{2}{*}{$\mathcal{L}_{ent}$} &
\multicolumn{3}{c}{BLIP-VQA} \\
& & & Color & Texture & Shape \\
\midrule
A & $\times$ & $\times$ & 0.5654 & 0.5815 & 0.5590 \\
B & $\checkmark$ & $\times$ & 0.6360 & 0.6139 & 0.5799 \\
C & $\times$ & $\checkmark$ & 0.6621 & 0.6274 & 0.5972 \\
\textit{\textbf{Ours}} & $\checkmark$ & $\checkmark$ & \textbf{0.7010} & \textbf{0.6971} & \textbf{0.6096} \\
\bottomrule
\end{tabular}
}
\label{tab:ablation_num}
\end{table}

\section{Conclusion}
In this paper, we conduct a comprehensive analysis of the semantic binding problem in T2I generation. We propose \textit{STEDiff}, a novel training-free method designed to alleviate semantic inconsistency issues that arise when models are prompted with complex textual descriptions. \textit{STEDiff} requires neither LLMs nor additional training. It activates text embeddings via [EOT] tokens by decomposing complex prompts into sub-sentences and replacing subject-related tokens, effectively reducing semantic inconsistency. Extensive experiments on T2I-CompBench validate its effectiveness over existing methods. The results demonstrate that, compared with prior approaches, \textit{STEDiff} is more effective at alleviating incorrect semantic binding and object disappearance issues that commonly arise in T2I models when handling complex prompts.

\noindent \textbf{Acknowledgments} This research is supported by Institute of Advanced Medicine and Frontier Technology (2023IHM01080); The computation is completed on the HPC Platform of Hefei University of Technology.

\bibliographystyle{ieeetr}
\bibliography{ref}
\end{document}